\documentclass[letterpaper, 10 pt, conference]{ieeeconf}  

\IEEEoverridecommandlockouts                              

\usepackage{graphics} 
\usepackage{epsfig} 

\usepackage{subfigure} 
\usepackage{stfloats} 
\usepackage{amsmath} 
\usepackage{flushend} 
\usepackage{cite} 
\usepackage{url} 
\usepackage{float} 
\usepackage{refstyle}

\newcommand{\reffig}[1]{Fig.~\ref{#1}}
\newcommand{\reftab}[1]{Table~\ref{#1}}
\newcommand{\refeqn}[1]{(\ref{#1})} 

\title{\LARGE \bf
Spatiotemporal Learning of Dynamic Gestures from 3D Point Cloud Data}

\author{Joshua Owoyemi and Koichi Hashimoto
\thanks{Authors are with the Department of System Information Sciences, Graduate School of Information Sciences, Tohoku University, Aoba-ku, Sendai 980-8579, Japan
        {\tt\small (email: tjosh@dc.tohoku.ac.jp; koichi@tohoku.ac.jp).}}%
}

\begin{document}

\maketitle
\thispagestyle{empty}
\pagestyle{empty}

\begin{abstract}
In this paper, we demonstrate an end-to-end spatiotemporal gesture learning approach for 3D point cloud data using a new gestures dataset of point clouds acquired from a 3D sensor. Nine classes of gestures were learned from gestures sample data. We mapped point cloud data into dense occupancy grids, then time steps of the occupancy grids are used as inputs into a 3D convolutional neural network which learns the spatiotemporal features in the data without explicit modeling of gesture dynamics. We also introduced a 3D region of interest jittering approach for point cloud data augmentation. This resulted in an increased classification accuracy of up to 10\% when the augmented data is added to the original training data. The developed model is able to classify gestures from the dataset with 84.44\% accuracy.
We propose that point cloud data will be a more viable data type for scene understanding and motion recognition, as 3D sensors become ubiquitous in years to come.

\end{abstract}

\section{INTRODUCTION}
In recent years, understanding human motion has been gaining more popularity \cite{Ye2013} in robotics, mainly for human-robot interaction (HRI)\cite{Nikolaidis2014} applications. This knowledge is useful because of increase in situations where humans and robots continue to share working and inhabiting spaces. There is therefore need for robots to `understand' the intentions of humans through gesture and action recognition, interpretation and prediction of human motions. This will help to effectively work with humans and ensure the safety of both parties.
While there has been a lot of research using 2D sensors and images for this purpose \cite{simonyan2014two} \cite{Wang2017}, 3D sensors have gained traction because of increased accessibility to low cost 3D sensors such as the Microsoft Kinect \cite{Marin2014}, allowing for more uses for 3D data. 

In this paper we are interested in learning human action and gestures from 3D point cloud data. According to \cite{Asadi-Aghbolaghi2017}, actions are more generic whole body movements while gestures more fine-grained upper body movements performed by a user that have a meaning in a particular context. In our previous work \cite{Owoyemi2017}, we have developed a model to predict the intentions of human arm motions in a workspace. We further evaluate this approach for dynamic gestures in this paper.

Researchers have used 2D data to achieved some remarkable progress in action and gesture recognition \cite{Ji2013} \cite{Lea2016} \cite{LeaCol2016} \cite{simonyan2014two}. However, using 2D images have proven difficult in some situations such as varying illumination conditions and cluttered backgrounds \cite{Rautaray2012}. 
On the other hand, 3D data such as point cloud and depth maps offer advantages such as illumination invariance \cite{Feng2017} and can capture appropriate information about the exact size and shape of an object in its physical space, allowing precise and accurate data usage for 3D manipulation, coordination and visualization \cite{Song2014} \cite{Apostol2014}. 

In most of the reviewed works, researchers used depth maps \cite{Wang2015} or skeleton representations \cite{Wu2012} \cite{Ding2017} from 3D sensors to analyze or learn human motion. To our knowledge, there has been few works in gesture recognition based on only 3D point cloud data. This might be largely because point cloud data are unorganized, making it difficult to be used directly for model inputs. Therefore, some considerable preprocessing has to be done in order to convert the raw data into usable formats. In our case, we make use of an occupancy grid representation, with unit cells referred to as voxels. This representation can be used for arbitrary size of point cloud spaces, subject to the resolution of the voxels.
The key contributions of our work are as follows: 
1. We demonstrate spatiotemporal learning from point cloud data through a new dataset of common Japanese gestures.
2. We develop a 3D CNN model which learns gestures end-to-end from 3D representation of point clouds stream and outputs a corresponding gesture class performed by the human.
3. We evaluate the 3D CNN model on the new dataset of common Japanese gestures.


\section{Related Work} 
With 3D sensors becoming ubiquitous in computer vision and robotic applications, the task of motion and gesture recognition from 3D data is one of increasing practical relevance. A general approach to gesture recognition from 3D data involves feature extraction from data, followed by a possible dimensionality reduction, and application of a classifier on the resulting processed data. 
Some of earlier approaches include; Hidden Markov Model (HMM) based classifier \cite{Cholewa2014} using the size of an ellipsoid containing an object and the length of vector from the center of mass to extreme points, analysis of Fourier transform and Radon transform of self-similarity matrix of features obtained from actions \cite{AsadiAghbolaghi2014}, and computation of local spin image descriptors \cite{Apostol2014} or Local Surface Normal (LSN) based descriptors\cite{Abdolmaleki2013}. A different approach was the use of multi-viewed projection of point cloud into view images and describing hand gestures by extracting and fusing features in the view images \cite{Liang2016}, claiming that conversion of feature space increases the inner-class similarity and reduces inter-class similarity. Converse to these aforementioned approaches, our approach does not rely on hand-engineered features or descriptors from point cloud data. Rather, we use a 3D convolutional neural network (3D CNN) based model which is able to both automatically build relevant features and classify inputs from example data.

\subsection{Other 3D Data Types}
Apart from point clouds, 3D data are also represented as depth maps, which are images that contains information relating to the distance of the surfaces of scene objects from a viewpoint. Depth maps have proven useful in gesture recognition \cite{Wang2017} \cite{Feng2016} \cite{Imran2016} mostly because the data is in 2D, which makes it easy to apply popular feature extraction approaches. However, we argue that our approach is applicable to not only gesture learning but action learning in general. While depth data offers only a point of view of the 3D space, using raw point cloud data allows us to define arbitrary regions of interest (ROIs) in the case of using sensors that provide 3D field of view. An example will be LIDAR sensor mounted on a self-driving car. We can create multiple ROIs depending on the location of objects in the scene and individually analyze their actions.

Secondarily, researchers also use skeletal features, provided by the sensor's software development kit (SDK) as in the case of Kinect\footnote{https://msdn.microsoft.com/en-us/library/dn799271.aspx}, or extracted manually from depth data\cite{Wu2012}. These methods \cite{Ding2017}\cite{Ding2017}\cite{Li2017}, however, involve manually modeling or engineering the features for the gestures or actions to be learned. Our approach, on the other hand, does not involve explicitly modeling the dynamics of the gestures or actions learned. Instead, the gestures are learned end-to-end, from input directly to gesture class using supervised learning.

\subsection{Gesture and action recognition with CNN}
CNNs have proven effective in pattern recognition tasks\cite{LeCun1998} and is known to outperform hand-engineered feature-based approaches in image classification, object recognition and similar tasks\cite{Krizhevsky2012}\cite{Wang2017}\cite{simonyan2014two}. For gesture recognition, CNNs have been used to achieve state-of-the-art results \cite{Ji2013}\cite{Ding2017}\cite{Lea2016}\cite{Li2017} utilizing different kind of feature representations and data types. Asadi-Aghbolaghi et al. \cite{Asadi-Aghbolaghi2017} presents a good survey on deep learning for action recognition in image sequences.

\begin{figure}[!ht]
\centering
\includegraphics{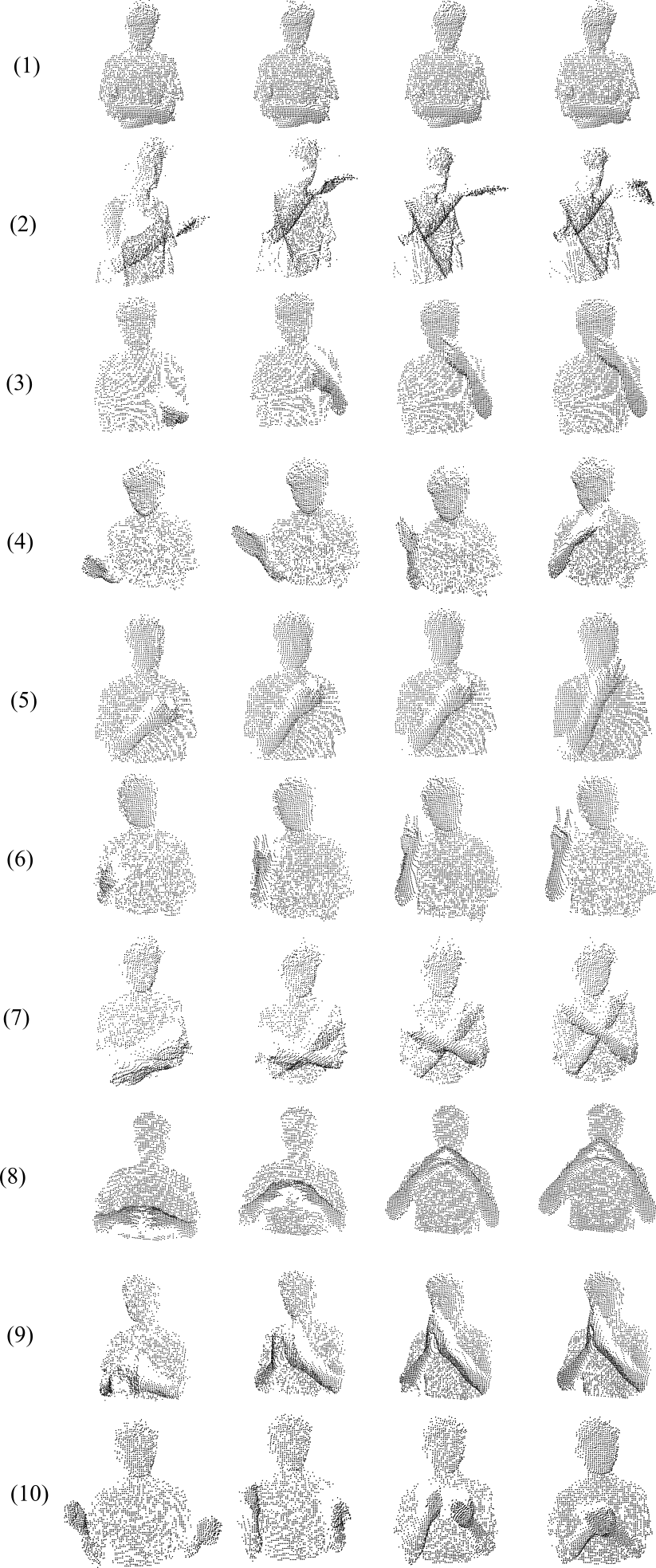}
\caption{
Sample frames of the Dataset of Common Japanese Gestures. We collected one class of no gestures and 9 classes of gestures. The gesture classes are (1) No gesture, (2) Come here, (3) Me, (4) No thank you, (5) Money, (6) Peace, (7) Not allowed, (8) OK, (9) I'm sorry, (10) I got it!. 
}
\label{gestures_array}
\end{figure}

In this paper, our model is inspired by the early fusion model in \cite{Karpathy2014} where consecutive frames in a video are fed into a 2D CNN in order to classify the actions in the video stream. Similarly, we also feed consecutive frames of point cloud data, however, into a 3D CNN to learn the actions performed in the point cloud stream. To demonstrate the spatiotemporal learning capability of our approach, we collected a new dataset of common Japanese gestures (See \reffig{gestures_array}). These were chosen arbitrarily from seeing videos and asking randomly chosen Japanese people to tell us the common gestures they use in a day-to-day life.

\section{METHOD}
In this section we describe the methods we used in learning spatiotemporal features from point cloud data. First, we present the data collected, then the training data preparation. We also describe the data augmentation approach employed to improve the efficiency of the learning process.
\subsection{Data collection} 
\begin{figure}[t]
\centering
\includegraphics{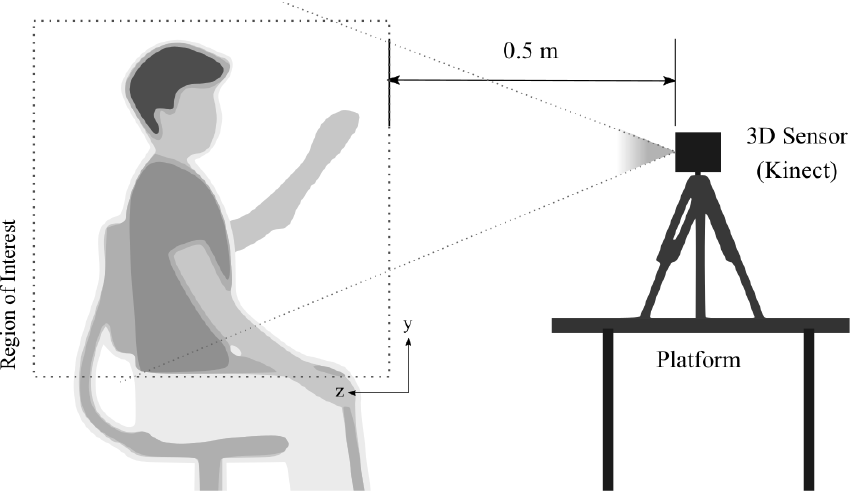}
\caption{
The setup for point cloud data collection. The subject sits in front of a Kinect sensor at a considerable distance. 
The ROI is -0.50 m to  0.50 m, -0.30 m to 0.40 m, 0.50 m to 1.40 m in x, y and z axes respectively.
}
\label{experiment_setup}
\end{figure}

We collected training data as point cloud frames acquired from a Kinect sensor. 
The dataset consist of a class labeled `No Gesture' and nine other gesture classes making a total of ten classes. The classes of gestures are: 1. No Gesture, 2. Come, 3. Me, 4. No Thank You, 5. Money, 6. Peace, 7. Not Allowed, 8. OK, 9. I'm Sorry, and 10. I Got It!. 
The dataset were collected with a Kinect Sensor facing subjects as shown in  \reffig{experiment_setup}. There were 5 subjects with each subject repeating the gestures at least 30 times per gesture. A total of 87,156 point cloud frames were collected for training and 29,758 data frames for testing. Sample frames of the dataset are shown in \reffig{gestures_array} for four consecutive frames of each gesture class. A ROI was determined to specify the volume in the scene where the subject is expected to be in. Therefore, the sensor only captures the upper part of the subject. The body was not removed from the point cloud because some gestures also involve bodily movements and the use of both hands. An example is the gesture "I'm sorry", which involves slightly bending or bowing the head with both hands clapped together in front of the head.

\begin{figure*}[hb!]
\centering
\includegraphics{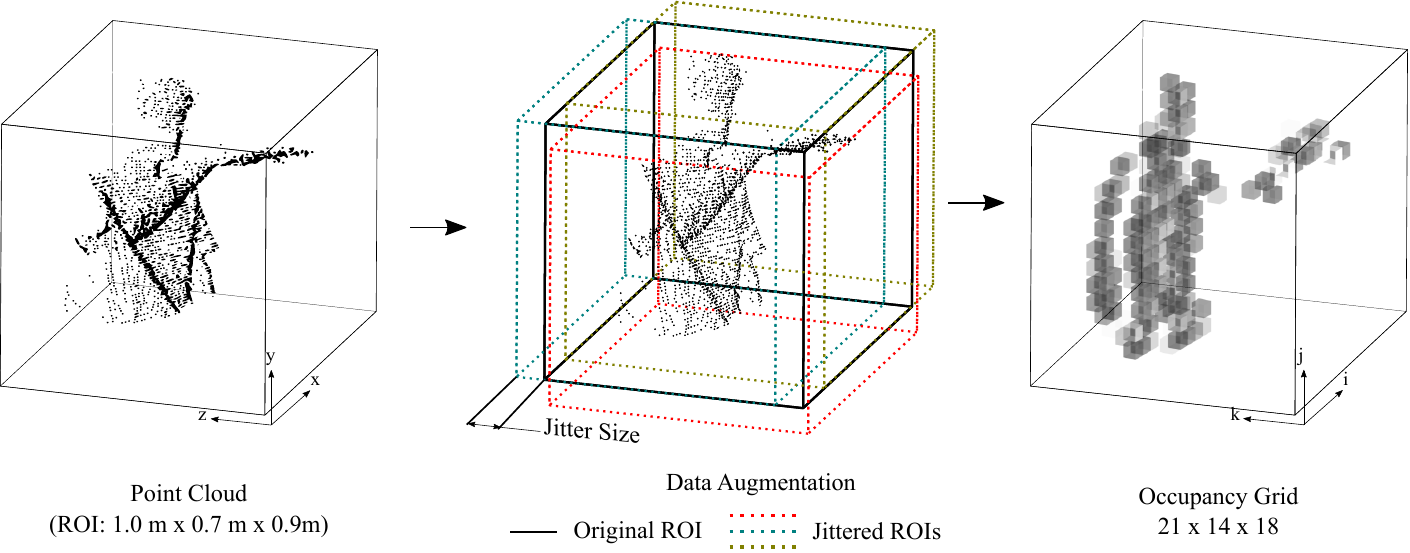}
\caption{
Input occupancy grid, mapped from point clouds. The dimension of the occupancy grid depends on the size chosen for the voxels. Here we used a voxel size of 50 mm. There is also a data augmentation step which involves randomly jittering the ROI by a size \(\alpha\) across a combination of the axes. To augment a point cloud data, the original ROI is jittered randomly in the x-axis or the y-axis or the z-axis or a combination of the three axes.
}
\label{ocg_conversion}
\end{figure*}



\subsection{Training Data Preparation}
To prepare the training data, point clouds from a 3D sensor are converted into 3D occupancy grids, with each point \((x,y,z)\) in the point cloud discretely mapped to a voxel coordinate \((i,j,k)\) by updating the occupancy grid similar to \cite{Maturana2015}. Each voxel has an initial value \(V_{ijk}^p=0\) and is updated by;
        
\begin{equation}
\label{eqn1}
V_{ijk}^p=V_{ijk}^{p-1}+z^p
\end{equation}

where \(\{z^p\}_{p=1}^P\) is a sequence of range measurement that either hit \(z^p=1\), or pass through \(z^p=0\) a given voxel and \(p\) is the individual points in the point cloud. This process is illustrated in \reffig{ocg_conversion} alongside the data augmentation approach which we will explain in the next subsection. We also define a ``lookback-window'' size \(m\) corresponding to the number of prior time steps to consider when recognizing the gesture at time \(t\). Hence, for each input time step, we have a data sample tensor \(\boldsymbol{V}_t=\big\{V_t, V_{t-1}, V_{t-2}, ... , V_{t-m+1}\big\}\) paired with the label \(Y_t\), the corresponding gesture class. Therefore we aim to find a set of parameters of the non-linear function that maps the input \(\boldsymbol{V}_t\) to the corresponding label \(Y_t\), given by \refeqn{eqn2}.

\begin{equation}
\label{eqn2}
Y_t = f\big\{\boldsymbol{V}_t\big\}
\end{equation}

\subsection{Training data augmentation}
We doubled the amount of training data by carrying out the data augmentation approach we call `3D ROI jittering' on the original dataset. This was achieved by applying a translation vector \(\begin{bmatrix} \alpha_{x}, \alpha_{y}, \alpha_{z}\end{bmatrix}^T\) to the original 3D ROI of each gesture performance in the dataset. As illustrated in \reffig{ocg_conversion}, basically we are shifting the ROI around in space while the position of the point cloud data is fixed. This is intended to simulate spatial variations in the performance of the gestures.

For an ROI, \(([x_{1}, x_{2}],[y_{1}, y_{2}],[z_{1}, z_{2}])\), and jittering size \(\alpha\), the jittered ROI can be obtained as: 

\begin{equation}
\begin{bmatrix}
x_{i,aug} \\
y_{i,aug} \\
z_{i,aug} \\
1
\end{bmatrix}_{a}
=
\begin{bmatrix}
    1 & 0 & 0 & \alpha_{x} \\
    0 & 1 & 0 & \alpha_{y} \\
    0 & 0 & 1 & \alpha_{z} \\
    0 & 0 & 0 & 1
\end{bmatrix}
\begin{bmatrix}
x_{i} \\
y_{i} \\
z_{i} \\
1
\end{bmatrix}
\end{equation}

Where the vector \(\begin{bmatrix} \alpha_{x}, \alpha_{y}, \alpha_{z}\end{bmatrix}^T\) is chosen randomly from the set,
\[
\bigg( 
\begin{tabular}{ccc}
0, & \(\alpha\) \\ 
\(\alpha_{x}\), & \(\alpha_{y}\), & \(\alpha_{z}\) 
\end{tabular} 
\bigg)
\cup
\bigg( 
\begin{tabular}{ccc}
0, & \( -\alpha\) \\ 
\(\alpha_{x}\), & \(\alpha_{y}\), & \(\alpha_{z}\) 
\end{tabular} 
\bigg), 
\]
the union of permutations with replacement for \(\{0, \alpha\}\) and \(\{0, -\alpha\}\). These permutations, therefore, represent the different jittering configurations we can have for a chosen jitter size.

\subsection{3D CNN Model}
CNN models are known to be suited for representing the non-linear relationships, as in \refeqn{eqn2}, involving a multidimensional input spaces such as image classification and object detection problems\cite{LeCun1998}. Here, we are interested in using a CNN model to learn gestures from 3D data, hence the use of 3D CNN. 

\begin{figure}[t!]
\centering
\includegraphics[scale=0.8]{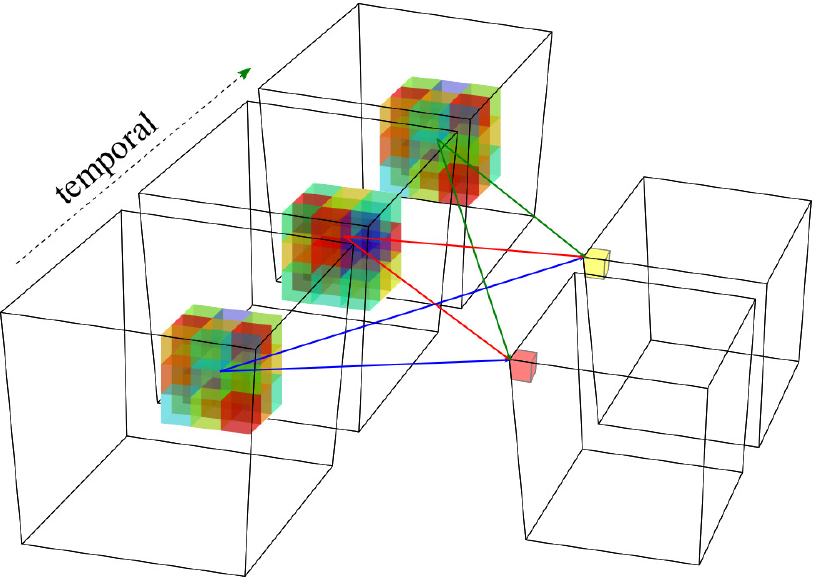}
\caption{
3D convolution illustration. 3D kernel, represented by the multicoloured cubes, are applied to inputs from previous layer. Here we also have a temporal dimension and the kernel weights are shared across this dimension. For the input layer, contiguous occupancy grid time steps represent the temporal dimension of the input.}
\label{convolution_illustration}
\end{figure}

CNNs are characterized by convolution operations between an input tensor \(\boldsymbol{I}\) and a convolution kernel \(\boldsymbol{K}\) (see illustration in \reffig{convolution_illustration}). For 3D inputs we have an output;

\begin{equation}
\label{convolution_eqn}
\begin{aligned}
A(i,j,k) & = (\boldsymbol{K}*\boldsymbol{I}) \\ 
         & = \sum_{l}\sum_{m}\sum_{n} \boldsymbol{I}(i-l, j-m, k-n)\boldsymbol{K}(l,m,n)
\end{aligned}
\end{equation}

which is the activation of the node \((i, j, k)\) of the feature map in the next layer.

Deep CNN models achieve automatic feature construction by stacking multiple convolutional layers, where higher layers capture more complex or discriminating features. Formally, each layer's output of the model is a set of activations \(\boldsymbol{A}\) from the layer's Rectified Linear Units (ReLUs) \cite{Maas2013} which are functions of kernel weights \(\boldsymbol{W}\), and biases \(\boldsymbol{b}\) to be optimized. Equations \refeqn{eqn3} to \refeqn{eqn5} represent operations at the layers. \(l\) are intermediate layers while \(L\) is the output layer.

\begin{equation}
\label{eqn3}
\boldsymbol{A}^0 = \boldsymbol{V}_t
\end{equation}
\begin{equation}
\label{eqn4}
\boldsymbol{A}^l = \boldsymbol{ReLU}(\boldsymbol{W}^l\boldsymbol{A}^{l-1}+\boldsymbol{b}^l)
\end{equation}
\begin{equation}
\label{eqn5}
\hat{Y} = \boldsymbol{Softmax}(\boldsymbol{W}^L\boldsymbol{A}^{L-1}+\boldsymbol{b}^L)
\end{equation}

\begin{figure*}[!ht]
\centering
\includegraphics[scale=0.9]{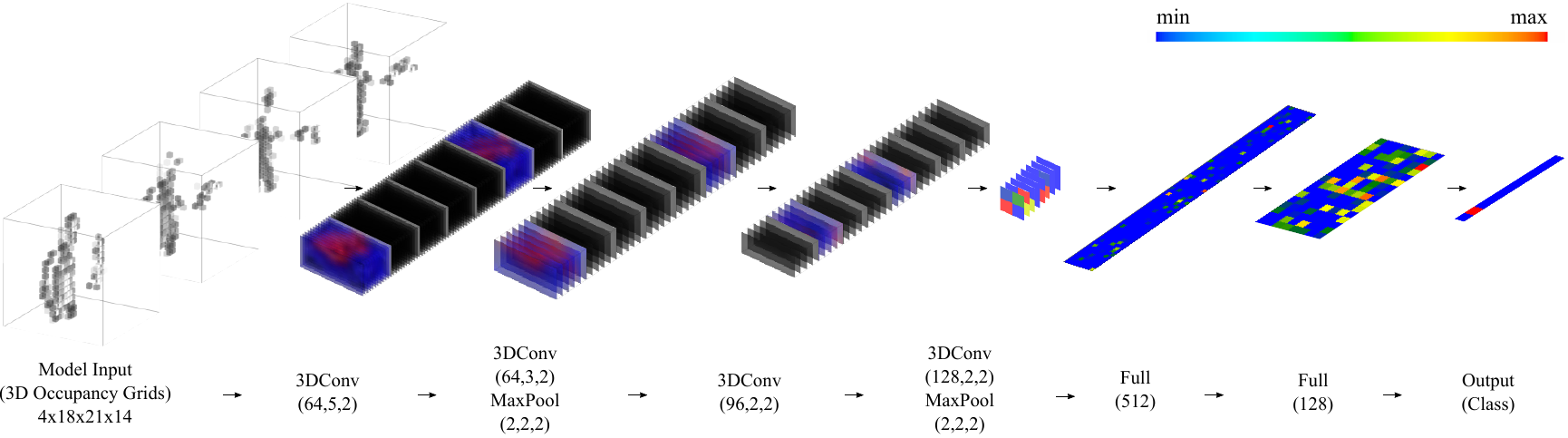}
\caption{
The architecture of the developed 3D CNN model. Input into the model are time steps of occupancy grids converted from point cloud data. Here, we show the first four filters of the convolutional layers and the activation map in the fully connected layers. The red cells show the activations in the filters of each layer. The filters that are black signify no activation for the given input.
}
\label{fig_full_arch}
\end{figure*}

Using similar configurations from previous work \cite{Owoyemi2017}, with modifications relating to the size of voxel space. Our CNN model configuration is as follows: 
As shown in \reffig{fig_full_arch}, we have a total of 7 layers, that is, 4 convolutional layers, 2 fully connected layers and the output layer. The 3D convolutional layers were designed to extract spatial features in each input time step and temporal features across time steps of the input. The first and second layers made use of 5x5x5 convolution kernels, followed by a 3x3x3 in the third layer and a 2x2x2 kernel in the fourth layer. A stride of 2x2x2 was used throughout the convolutional layers. We apply max pooling \cite{scherer2010evaluation} on the second and fourth layer to reduce the dimensionality of the parameters connected to the next layers.

\subsection{Training Details}
We trained our final model using a 5-fold cross-validation, employing an early stop approach of a patience of 3 in each cross-validation cycle. This prevents the model from overfitting and helps to choose a good trade-off between accuracy and training loss. For optimization, we used an Adam optimizer \cite{Kingma2015}, and used dropouts \cite{Hinton2012} of 0.3 on the fully connected layers to further prevent overfitting. During training, we reduce the learning rate by a factor of 0.3 after the validation loss has not decreased in 3 epochs.

All training was done on an Intel(R) Xenon(R) CPU E5-2637 v4 @ 3.50GHz x 12 with a 4 NVIDIA TITAN X GPUs. The Keras\footnote{https://keras.io/} library with Tensorflow\footnote{https://www.tensorflow.org/} backend was used for model implementation.

\section{EVALUATION}
The developed model was evaluated on the test data kept apart during data collection. We used a window size of 4 timesteps for our training and evaluation. The following subsections outline details of the evaluations carried out on the model.
\subsection{Data Augmentation}
\begin{figure}[t] 
\centering
\includegraphics[scale=0.9]{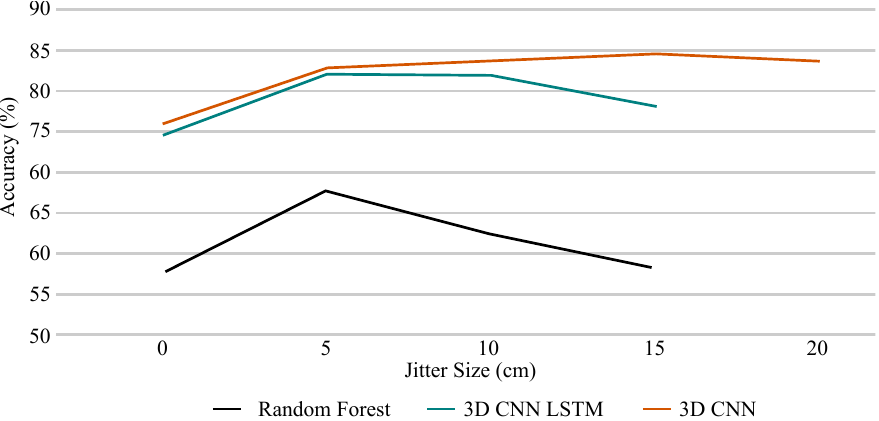}
\caption{
Evaluation and comparison of augmentation jitter sizes. Using jitter sizes of 5cm, 10 cm and 1.5 cm. The jitter size of 0 cm signifies no augmentation was performed. It is observed that the accuracies of both the  3D CNN and the off-the-shelf classifier increased after applying data augmentation. The accuracies increase, from 57.4\% to 67.64\% for the the random forest model, from 74.47\% to 81.98\% for the 3D CNN LTSM model and from 75.8\% to 82.8\% for the 3D CNN model. However a further increase in the jitter size did not yield significant increase in accuracy. It rather becomes worse, more evidently in the random forest model and later for other models.
}
\label{jitter_evaluation}
\end{figure}

Training with augmented data not only increased the number of training data that was used by 100\% but also helped in compensating for the spatial variation in the gestures since the subject could perform the gesture in different positions within the ROI. We compared the result of training with and without augmented data and found that adding augmented data to our training samples increased test accuracy up to 10\%. This is true for the different classifiers that we evaluated. This is a significant improvement from using only the collected data. A summary of the results from data augmentation is shown in \reffig{jitter_evaluation}, showing the comparison of different jitter sizes on different classifiers.

\subsection{Models Comparison}
\begin{figure}[t]
\centering
\includegraphics{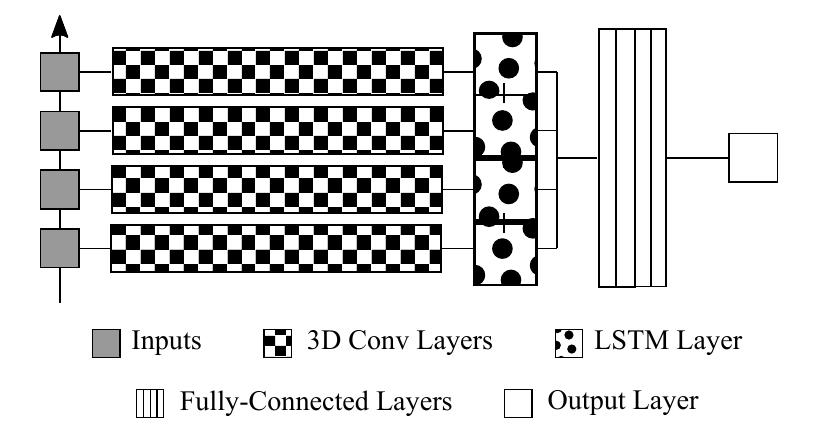}
\caption{
The architecture of the LSTM variant of 3D CNN of our final model. The input time steps are separated into individual mini CNN networks in order that the spatial features are learned in the CNN layers and then the temporal features in the LSTM layers. This achieved a classification accuracy of 81.82\%.
}
\label{3dcnnlstm_arch}
\end{figure}

We compared the 3D CNN model with a random forest model, an off-the-shelf classifier. This comparison is to evaluate if the 3D CNN model has considerable accuracy advantage over an off-the-shelf classifier. On the other hand, LSTM models \cite{Greff2016} are known to perform better on time series or temporal data, so we compared the 3D CNN model with an LSTM variant. This was done by replacing the first fully connected layers in the model with an LSTM layer and passing each time steps of the input through individual mini 3D CNN networks to learn spatial features, and then into the LSTM layers to learn the temporal features. The LSTM variant of the 3D CNN architecture is shown in \reffig{3dcnnlstm_arch}. Even though these two models are not directly equivalent, we used similar hyperparameters for training in other to have similar conditions. In our evaluation, the 3D CNN LSTM model did not perform better than the 3D CNN. A summary of this evaluation results is shown in \reftab{tbl: performance_table}

\begin{figure}[t]
\centering
\includegraphics{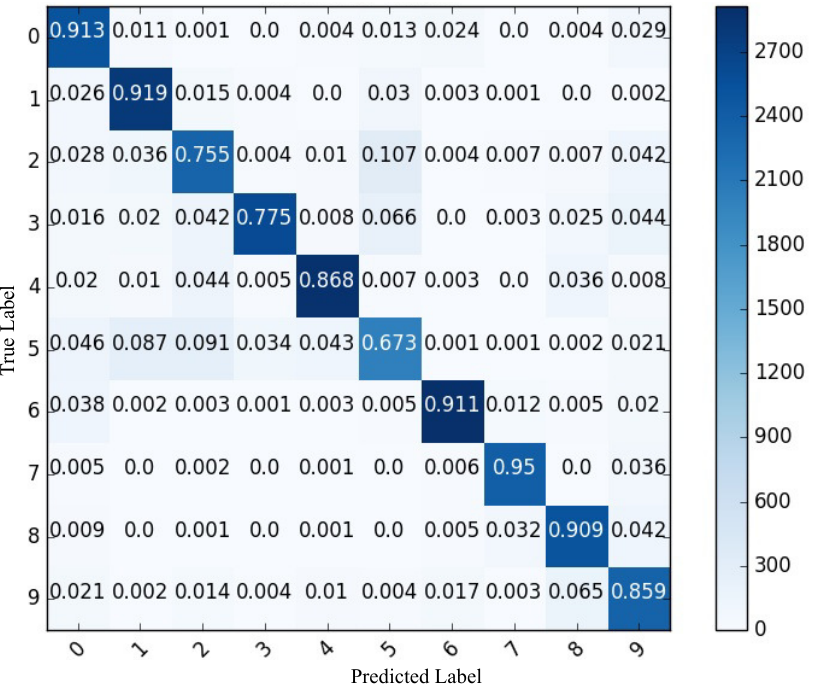}
\caption{
Confusion Matrix of the prediction on the test set, showing the performance of the model in each class. Class 5 gesture, Peace, has the lowest accuracy, while class 6, Not Allowed, has the highest accuracy.
}
\label{confusion_matrix}
\end{figure}

We see that the 3D CNN model is able to learn both spatial and temporal relationship in the data presented, furthermore, it outperforms the LSTM variant of the model for this particular problem.
The confusion matrix for our final model is shown in \reffig{confusion_matrix}.

\begin{table}[t]
\begin{center}
 \caption{Models Performance Comparison}
 \label{tbl: performance_table}
 \begin{tabular}{|c|c|c|c|}
  \hline
        \bfseries Model & \bfseries Accuracy \\ \hline \hline
        Random Forest & 67.64\% \\ \hline
        3D CNN  & 75.80\% \\ \hline
        3D CNN + Augmentation & \bfseries 84.44\% \\ \hline
        3D CNN + LSTM & 74.47\% \\\hline
        3D CNN + LSTM + Augmentation & 81.82\% \\ \hline
  \hline
 \end{tabular}
 \end{center}
\end{table}

\section{CONCLUSIONS}
We showed an end-to-end approach for spatiotemporal gesture learning from point cloud data. Our data augmentation approached achieved an increase in the model accuracy up to 10\%. One limitation for this work is the ability to work with higher resolution representation of point clouds. A smaller voxel size would dramatically increase the dimension of the training data, and subsequently the training and computation time involved. At some point a smaller voxel size in infeasible because of limited computation resource and memory. A future work could address an approach for computation and memory efficient representation of point clouds, or other data augmentation scheme that compensates for anthropometry of subjects used for training thereby covering a wide range of users.

\section*{ACKNOWLEDGMENT}
This work is partially supported by JSPS Grant-in-Aid
16H06536.

\bibliography{ms}
\bibliographystyle{IEEEtran.bst}



\end{document}